# Chain graphs for learning

Wray L. Buntine*
RIACS at NASA Ames Research Center
Mail Stop 269–2
Moffett Field, CA 94035–1000, USA
wray@kronos.arc.nasa.gov

## Abstract

Chain graphs combine directed and undirected graphs and their underlying mathematics combines properties of the two. This paper gives a simplified definition of chain graphs based on a hierarchical combination of Bayesian (directed) and Markov (undirected) networks. Examples of a chain graph are multivariate feed-forward networks, clustering with conditional interaction between variables, and forms of Bayes classifiers. Chain graphs are then extended using the notation of plates so that samples and data analysis problems can be represented in a graphical model as well. Implications for learning are discussed in the conclusion.

## 1 Introduction

Probabilistic networks are a notational device that allow one to abstract forms of probabilistic reasoning without getting lost in the mathematical detail of the underlying equations. They offer a framework whereby many forms of probabilistic reasoning can be combined and performed on probabilistic models without careful hand programming. Efforts to date have largely focused on first-order probabilistic inference, for instance found in expert systems and diagnosis (Heckerman, Mamdani, & Wellman, 1995; Spiegelhalter, Dawid, Lauritzen, & Cowell, 1993), and planning (Dean & Wellman, 1991). For instance, given a set of observations about a patient, what are the posterior probabilities for different diseases? Should an additional expensive diagnostic test be performed on the patient? This paper presents a representation for extending probabilistic networks to handle second-order or statistical problems. Second order problems are mainly concerned with building or improving a probabilistic network from a database of cases. Second order inference on probabilistic networks was first suggested by Lauritzen and Spiegelhalter (Lauritzen & Spiegelhalter, 1988), and has subsequently been developed by several groups (Gilks, Thomas, & Spiegelhalter, 1993; Dawid & Lauritzen, 1993; Buntine, 1994; Shachter, Eddy, & Hasselblad, 1990). Whereas, an introduction to learning of Bayesian networks can be found in (Heckerman, 1995).

This paper uses *chain graphs* (Lauritzen & Wermuth, 1989) as a general probabilistic network model. Chain graphs mix undirected and directed graphs (or networks) to give a probabilistic representation that includes Markov random fields and various Markov models. Lauritzen and Wermuth demonstrated that chain graphs are a powerful tool for modeling statistical analysis, research hypotheses, and hence learning (Wermuth & Lauritzen, 1989). Chain graphs when augmented with deterministic nodes can represent many well known models as a special case including generalized linear models, various forms of clustering, feed-forward neural networks and various stochastic neural networks. This includes a large number of the more general network models now available (Ripley, 1994). These many different models are formed by combining basic nodes in the network representing for instance, Gaussian variables or deterministic Sigmoid units. The expressiveness of chain graphs is illustrated in Section 6 where a number of models are represented. Decision theoretic constructs could also be used to represent the decisions and utilities of a problem (Shachter, 1986), although this is not done here.

In this paper, I define a chain graph as a hierarchical combination of directed (Bayesian) and undirected (Markov) networks. This definition extends the notion of block recursive models used in (Wermuth & Lauritzen, 1989; Højsgaard & Thiesson, 1995) and analyzed in (Frydenberg, 1990, Theorem 4.1) by allowing blocks to include directed networks as well as undirected networks. This definition allows the complex independence properties and functional form of a chain graph (Frydenberg, 1990) to be read off from knowledge of the simpler corresponding properties for directed and undirected networks. This definition also allows chain graphs to have embedded deterministic nodes—as commonly used in learning for neural networks, generalized linear models, and basis functions—and thus extends the general applicability of chain graphs. Previous constructions relied on positivity constraints (Frydenberg, 1990) so did not allow determinism, or used limit theorems (Højsgaard & Thiesson, 1995) to allow some determinism. This framework for modeling chain graphs, and the general interpretation theorem, Theorem 2, are the major technical contribution of this paper.

---
*Current address: Heuristicrats Research, Inc., 1678 Shattuck Avenue, Suite 310, Berkeley, CA 94709-1631, wray@Heuristicrat.COM

First, Sections 2 and 3 review basic results on directed and undirected networks, as for instance introduced in (Pearl, 1988; Whittaker, 1990). Necessary independence properties and functional representations of these networks, as needed for chain graphs, are summarized. Then the notion of conditional networks are formalized in Section 4. Conditional networks are implicitly used throughout the community, however we need to define them carefully as a building block for the definition of chain graphs. A chain graph is then introduced as a hierarchical combination of conditional networks, in Section 5. Several examples of learning systems are then given. Finally, a graphical construct for modeling samples and learning, plates (Buntine, 1994), is reviewed.

## 2   Directed networks

A Bayesian or directed network consists of a set of variables $X$ and a directed graph defined on it consisting of a node corresponding to each variable and a set of directed arcs. Nodes in the graph and the variables they represent are used interchangeably. The graph is such that it contains no directed cycles. In this paper, a directed network defines a particular functional form for the probability distribution $p(X)$ over the variables. Each variable is written conditioned on its *parents*, where $parents(x)$ is the set of variables with a directed arc into $x$. The general form for this equation for a set of variables $X$ is:

$$p(X) = \prod_{x \in X} p(x|parents(x)) . \qquad (1)$$

This functional form is the *interpretation* of a directed network used in this paper. The lemma below shows that this definition is equivalent to a definition based in independence statements (Lauritzen, Dawid, Larsen, & Leimer, 1990), related to (Pearl, 1988). The independence notation is due to (Dawid, 1979).

**Definition 1** *A is independent of B given C, denoted $A \perp\!\!\!\perp B | C$, when $p(A \cup B | C) = p(A|C)p(B|C)$ for all instantiations of the variables $A, B, C$.*

The following definitions are used here.

**Definition 2** *The ancestral set, $ancestors(A)$, of a subset $A$ of variables $X$ is the transitive closure of the relation, $f(B) = B \cup parents(B)$. The moralized* graph $G^m$ of a directed graph $G$ is formed by connecting every two nodes that have a common child with an undirected arc, and then dropping the directions from all directed arcs.

The particular independence statements are based on set separation in the moralized graph, which is equivalent to another condition known as d-separation (Pearl, 1988):

**Definition 3** *The distribution $p(X)$ satisfies the directed global Markov property relative to the directed graph $G$ if $A \perp\!\!\!\perp B | S$ when $S$ separates $A$ and $B$ in the graph $H^m$ where $H$ is the subgraph of $G$ restricted to $ancestors(A \cup B \cup S)$.*

**Lemma 1** *Given a directed graph $G$ on $X$, and $A, B, S \in X$. The distribution $p(X)$ satisfies the directed global Markov property relative to $G$ if and only if Equation (1) holds.*

Given a directed graph, we can therefore read off both the functional decomposition of Equation (1) and the independence properties easily.

## 3   Undirected networks

Similarly, a Markov or undirected network is an undirected graph on a set of variables $X$ representing a probability distribution $p(X)$ over the variables. This is analogous to Lemma 1, except that $p(X)$ must now be strictly positive. The appropriate independence conditions are based on set separation without first moralizing the graph.

**Definition 4** *The distribution $p(X)$ satisfies the global Markov property relative to the undirected graph $G$ if $A \perp\!\!\!\perp B | S$ when $S$ separates $A$ and $B$ in the graph $G$.*

The neighbors for a node $x$, denoted $neighbors(x)$ are the set of variables directly connected by an undirected arc to $x$. An important concept is the set of cliques on the graph.

**Definition 5** *The set of maximal cliques on $G$ is denoted $Cliques(G) \subset 2^X$. $C \in Cliques(G)$ if and only if $C$ is fully connected in $G$ and no strict superset of $C$ is fully connected in $G$.*

**Theorem 1** *An undirected graph $G$ is on variables in the set $X$. The distribution $p(X)$ is strictly positive in the domain $\times_{x \in X} domain(x)$. Then the distribution $p(X)$ satisfies the global Markov property if and only if $p(X)$ satisfies the equation*

$$p(X) = \prod_{C \in Cliques(G)} f_C(C) , \qquad (2)$$

*for some functions $f_C > 0$.*

The proof follows directly from (Frydenberg, 1990; Buntine, 1994). A form of this theorem for finite discrete domains is called the Hammersley-Clifford Theorem (Geman, 1990; Besag, York, & Mollie, 1991). Again, Equation (2) is used as the interpretation of an undirected network.

## 4 Conditional networks

Networks can also represent conditional probability distributions. *Conditional networks* are represented by introducing shaded variables in the graph. Shaded variables are assumed to have their values known, so the probability defined by the network is now conditional on the shaded variables. Figure 1 shows two conditional versions of a simple medical problem (Shachter & Heckerman, 1987). If the shading of nodes is ignored, the joint probability, $p(Age, Occ, Clim, Dis, Symp)$ for the two graphs (a) and (b) respectively is:

(a)  $p(Age)\,p(Occ|Age)\,p(Clim|Age, Occ)$
  $\phantom{(a)}\;p(Dis|Age, Occ, Clim)\,p(Symp|Age, Dis)$ ,

(b)  $p(Age)\,p(Occ)\,p(Clim)\,p(Dis|Age, Occ, Clim)$
  $\phantom{(b)}\;p(Symp|Age, Dis)$

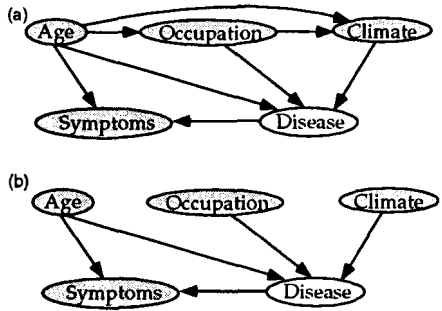

Figure 1: Two equivalent conditional models of the medical problem

However, because four of the five nodes are shaded, this means their values are known. The conditional distributions computed from the above are identical:

$p(Dis|Age, Occ, Clim, Symp)$

$$= \frac{p(Dis|Age, Occ, Clim)\,p(Symp|Age, Dis)}{\sum_{Dis} p(Dis|Age, Occ, Clim)\,p(Symp|Age, Dis)} \;.$$

More generally, conditional networks can be simplified sometimes (Buntine, 1994). The following simple lemma applies to conditional Bayesian networks and is derived directly from Equation (1).

**Lemma 2** *Given a directed network $G$ with some nodes shaded representing a conditional probability distribution. If a node $X$ and all its parents have their values given, then the Bayesian network $G'$ created by deleting all the arcs into $X$ represents an equivalent probability model to the Bayesian network $G$.*

A corresponding result holds for undirected graphs, and follows directly from Theorem 1.

**Lemma 3** *Given an undirected graph $G$ with some nodes shaded representing a conditional probability distribution. Delete an arc between given nodes $A$ and $B$ if all their common neighbors are given. The resultant graph $G'$ represents an equivalent probability model to the graph $G$.*

Furthermore, the probability formula for conditional networks follow easily as well from the corresponding Equations (1) and (2).

**Lemma 4** *Let $G$ be a conditional directed graph on variables $X \cup Y$ where variables $Y$ are given and $X$ are not. Then if $Y = ancestors(Y)$,*

$$p(X|Y) \;=\; \prod_{x \in X} p(x|parents(x)) \;. \qquad (3)$$

*If $G$ is a conditional undirected graph, then*

$$p(X|Y) \;=\; f_Y(Y) \prod_{C \in Cliques(G)} f_C(C) \;, \qquad (4)$$

*for some function $f_Y$ on $Y$.*

Again, Equations (3) and (4) are used as the interpretation of conditional undirected network, and the marginal distribution for the shaded variables, $p(Y)$, is ignored. Furthermore, independence statements derived from the graph are only valid if they are conditioned on the shaded variables $Y$.

## 5 Chain graphs

A chain graph is a graph consisting of mixed directed and undirected arcs, where any cycle with some directed arcs must contain at least two directed arcs in reverse direction. It is shown below that a chain graph can be represented as a hierarchical combination of conditional networks. A chain graph is first broken up into components as follows. The chain components are the standard definition used (Lauritzen & Wermuth, 1989).

**Definition 6** *Given a chain graph $G$ over some variables $X$. The chain components are the coarsest mutually exclusive and exhaustive partition of $X$ where the set of subgraphs induced by the partition are connected and undirected. Let chain-components$(A)$ denote all nodes in the same chain component as at least one variable in $A$.*

The chain components are unique and are found by removing the directed arcs from the graph $G$ and identifying the connected components on the resulting graph (Lauritzen & Wermuth, 1989).

**Definition 7** *Given a chain graph $G$ over some variables $X$. The component subgraphs are a coarser partition of variables $X$ than the chain components, and* *are the coarsest partition where the set of subgraphs induced by the partition are connected, undirected or directed (but not mixed) subgraphs of the chain graph G.*

These definitions imply that:

- a connected directed network has only a single component subgraph, itself; and
- likewise a connected undirected network has only a single component subgraph, itself.

This makes the component subgraphs a natural decomposition of the chain graph into its maximal directed and undirected parts. The following lemma shows the component subgraphs are unique.

**Lemma 5** *There is a unique set of component subgraphs for a chain graph G found by the following algorithm:*

> *Let U be the set of chain components for graph G, and let S be the set of singleton sets in U. Let D be the connected components in the graph $G_S$ (the graph G restricted to the variables in S, which is directed by construction). The component subgraphs are given by $(U - S) \cup D$ and the subgraphs they induce.*

**Proof** A component subgraph that is directed cannot contain two variables connected by an undirected arc. Furthermore, it is a superset of one or more chain components since component subgraphs are by definition a coarser partition than the chain components. Therefore, the component subgraphs that are directed must be formed by merging singleton chain components that are connected by directed arcs. This is the connectivity relation which has a unique coarsest partition. The algorithm above follows from this.  □

An example is given in Figure 2. Figure 2(a) shows the original chain graph. The chain components of G are $\{a, b\}$, $\{c\}$, $\{d\}$, and $\{e, f, g, h\}$. The component subgraphs are formed by merging c and d into a directed graph on c, d. Figure 2(b) shows on the left the three directed and undirected component subgraphs with their parents added and shaded. On the right is the Bayesian network showing how the component subgraphs are pieced together. Theorem 2 below shows this leads to the following functional form for the joint probability.

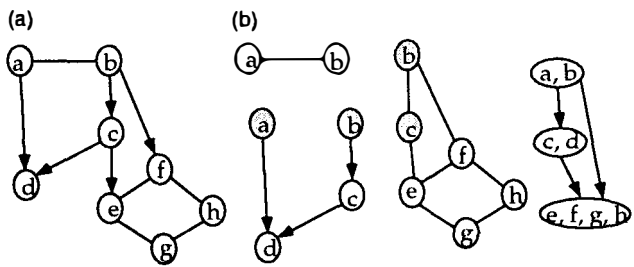

Figure 2: Decomposing a chain graph

$$p(a, b, c, d, e, f, g, h) =$$
$$p(a, b)\, p(c|b)\, p(d|a, c)$$
$$f_0(b, c) f_1(c, e) f_2(b, f) f_3(e, f) f_4(f, h) f_5(h, g) f_6(g, e) \;.$$

Notice that $f_0(b, c)$ exists to normalize $p(e, f, g, h|b, c)$.

Informally, a chain graph over variables $X$ with component subgraphs given by the set $T$ is interpreted first as the *component factorization* (compared to a block factorization (Højsgaard & Thiesson, 1995)) given by:

$$p(X) = \prod_{\tau \in T} p(\tau | parents(\tau)) \;, \qquad (5)$$

where

$$parents(A) = \bigcup_{a \in A} parents(a) - A \;,$$

and likewise for ancestors. The conditional probability $p(\tau | parents(\tau))$ for each component subgraph is now defined as a conditional directed or undirected network.

This can be formalized to give a definition for the interpretation of a chain graph. This works through the steps given for interpreting Figure 2.

**Definition 8** *Given a chain graph G on variables X with no given nodes. Let $U_1, \ldots, U_C$ be the component subgraphs of G. Construct a matching set of subgraphs $G_1, \ldots, G_C$ as follows. Let $G_i$ be the subgraph induced by G on $U_i \cup parents(U_i)$. Then, make the variables in $parents(U_i)$ all be shaded in $G_i$ and add extra arcs to make $parents(U_i)$ into a clique. The extra arcs should be directed if $U_i$ is a directed component subgraph, and undirected if $U_i$ is undirected. Now construct a directed graph $G_M$ whose nodes are $U_1, \ldots, U_C$ and arcs connect $U_i$ to $U_j$ if a variable in $U_i$ has a child in $U_j$ in the graph G. Then the chain graph G is defined to be equivalent to the set of subgraphs $G_1, \ldots, G_C$ together with the master graph $G_M$.*

The subgraphs $G_i$ can also be simplified according to Lemmas 2 and 3. One advantage of this formulation is that only undirected subcomponents of the chain graph need have the condition of positivity on their conditional distribution, required for Theorem 1 to hold. An additional example of Definition 8 appears in Figure 3. The original chain graph is in Figure 3(a). In this case, the chain components are $\{\{a, b\}, \{c, d\}, \{e\}, \{f\}\}$. The component subgraphs are $\{\{a, b\}, \{c, d\}, \{e, f\}\}$. The graphs $G_1, G_2, G_3$ are shown in Figure 3(b) along with the master graph $G_m$. The joint probability, whose general form is derived below, is

$$p(a, b, c, d, e, f) =$$
$$p(a, b)\, f_0(a, b)\, f_1(a, c)\, f_2(b, d)\, p(e|c)\, p(f|e, d)$$

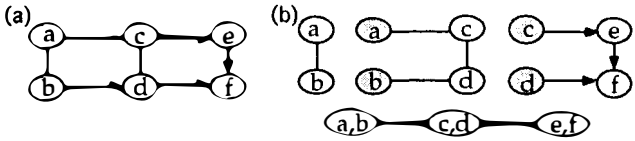

Figure 3: Decomposing another chain graph

The global Markov property for chain graphs is defined in the same way as the directed global Markov property. This requires that a chain graph be moralized. The definition below describes how this is done.

**Definition 9** *The ancestral set, ancestors(A), of a subset A of variables X is the transitive closure of the relation, $f(B) = B \cup neighbors(B) \cup parents(B)$. The moralized graph $G^m$ of a chain graph G is formed by connecting every two nodes that have children in a common chain component with an undirected arc, and then dropping the directions from all directed arcs.*

The corresponding relationship between independence and the functional form of the probability distribution then follows from Lemma 1 and Theorem 1, although not trivially so.

**Theorem 2** *A chain graph G is on variables in the set X. For every $U \subset X$ a chain component of G with cardinality greater than 1, the conditional distribution $p(U|parents(U))$ is strictly positive in the domain $\times_{x \in U} domain(x)$. Then the distribution $p(X)$ satisfies the global Markov property if and only if $p(X)$ satisfies Equations (3) and (4) for each of its subgraphs and master graphs.*

**Proof** The proof uses the notation from Definition 8. First, assume the joint probability $p(X)$ satisfies Equations (3) and (4) as required for the component subgraphs and the master graph, and prove the global Markov property holds. The joint probability therefore satisfies Equation (5) where $T$ is the set of component subgraphs and the terms $p(\tau|parents(\tau))$ are given by Equations (3) and (4). Since the directed component subgraphs satisfy Equations (3), it follows that

$$p(X) = \prod_{\tau \in T'} p(\tau|parents(\tau)) ,$$

where $T'$ are the chain components. Consider testing whether $A \perp\!\!\!\perp B \mid D$. Let $Z = ancestors(A \cup B \cup D)$. By construction, the marginal distribution for $Z$ is a restriction of this product to the chain components for $Z$.

$$p(Z) = \prod_{\tau \in (T' \cap Z)} p(\tau|parents(\tau)) .$$

where again the terms $p(\tau|parents(\tau))$ are given by Equation (4) for $\tau$ non-singleton. Let $G^m_Z$ be the moralized version of the chain graph $G$ restricted to $Z$. Since the moralized graph joins parents with an undi-

rected arc, $p(Z)$ takes the form of

$$p(Z) = \prod_{C \in Cliques(G^m_Z)} f_C(C) .$$

Now assume $D$ separates $A$ and $B$ in $Z$. The following shows $A \perp\!\!\!\perp B \mid D$. Due to separation, if we remove variables in $D$ from $G^m_Z$, the resulting graph separates into a part containing $A$, $Z_A$, and a part containing $B$, $Z_B$, where $A \cap Z_B = \emptyset = B \cap Z_A$. Ignoring variables in $D$, each clique in $G^m_Z$ must be wholly contained in one part or another. Hence, $p(Z) = f_1(Z_A, D) f_2(Z_B, D)$, and therefore independence holds by marginalizing out variables in $Z - A - B - D$.

Now for the reverse direction. Assume the global Markov property holds. Equation (5) follows directly. Now apply the global Markov property to each of the subgraphs $G_i$ in Definition 8. Suppose $G_i$ is directed. Let $x \subseteq U_i$. Since $x$ has no neighbors in $G$, it follows from the global Markov property for the chain graph that $x \perp\!\!\!\perp ancestors(x) \mid parents(x)$. Notice that ancestors in $G_i$ are a subset of the ancestors in $G$ so Equation (3) follows. Suppose $G_i$ is undirected. The condition in the theorem implies that $p(U_i|parents(U_i))$ is strictly positive in the domain $\times_{x \in U_i} domain(x)$. Also, the global Markov property implies that for $x \in U_i$, $x \perp\!\!\!\perp (U_i \cup parents(U_i)) \mid (neighbors(x) \cup parents(x))$. Using a proof like that for (Buntine, 1994, Theorem 2.1), Equation (4) follows.  □

A number of interesting properties can be derived from the global Markov property for chain graphs, their equivalence, and their relationship with Bayesian networks (Frydenberg, 1990; Andersson, Madigan, & Perlman, 1994). For instance, a chain graph is a convenient representation for the class of equivalent Bayesian networks (Verma & Pearl, 1990).

# 6    Examples of chain graphs for learning

This section presents a number of models using chain graphs, to illustrate their generality.

## 6.1    Feed-forward networks and deterministic nodes

The preceding definitions of a chain graph have been carefully set up to allow nodes to represent deterministic variables. Consider a feed-forward network, popular in neural networks, and general enough to represent logistic or linear regression. A simple feed-forward network is given in Figure 4(a). The corresponding chain graph is given in Figure 4(b), which also introduces a bivariate Gaussian error model on the output variables. Here the five sigmoid units of the network are modeled with deterministic nodes. A deterministic node has double circles to indicate it is a deterministic function of its inputs. The analysis of deterministic nodes in Bayesian networks and, more generally, in influence diagrams is considered by (Shachter, 1990). The network outputs $m_1$ and $m_2$ represent the mean of a bivariate Gaussian.

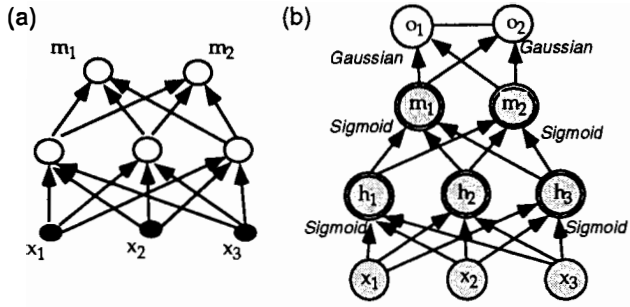

Figure 4: A feed-forward network and its chain graph

To analyze these nodes, we need to extend the usual definition of a parent and a child for a graph. Only one case is given here because it is all that is used in the lemma below.

**Definition 10** *The* non-deterministic children *of a node $x$ are the set of non-deterministic variables $y$ such that there exists a directed path from $x$ to $y$ given by $x, y_1, \ldots, y_n, y$, with all intermediate variables $(y_1, \ldots, y_n)$ being deterministic.*

For instance, in the model in Figure 4, the non-deterministic children of $x_2$ are $o_1$ and $o_2$. Deterministic nodes can be removed from a graph by rewriting the equations represented into the remaining variables of the graph. This goes as follows:

**Lemma 6** *A chain graph $G$ with nodes $X$ has deterministic nodes $Y \subset X$. The chain graph $G'$ is created by adding to $G$ a directed arc from every node to its non-deterministic children, and by deleting the deterministic nodes $Y$. The graphs $G$ and $G'$ are equivalent probability models on the nodes $X - Y$.*

An application of this lemma to the chain graph in Figure 4 is given in Figure 5. This of course, destroys

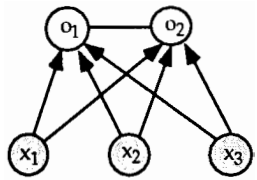

Figure 5: The non-deterministic version of a feed-forward network

all the information conveyed by the original graph.

## 6.2 Stochastic neural networks

Stochastic networks form the basis of the stochastic Boltzmann machine, and the Hopfield network (Hertz,

Krogh, & Palmer, 1991), which both have relationships to graphical models (Neal, 1992). A stochastic network corresponds to an undirected network with hidden variables, except interactions involve quadratic terms at most. A simple configuration is given in Figure 6. On the left is a representation of a network for a Boltzmann using the notation in (Hertz et al., 1991). The input variables are $x_1$, $x_2$, $x_3$ and $x_4$ and the output variable is $o$. There is one hidden variable $h_1$ marked in black. The corresponding chain graph is on the right with the parameters for the model, the weights, explicitly represented. The weights of the feed-forward network were not represented in the example of the previous section. This chain graph has four chain components.

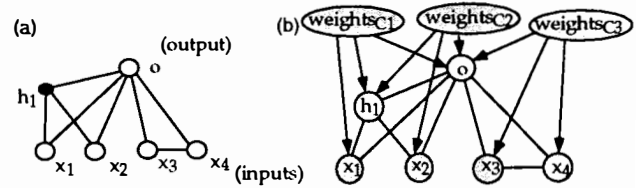

Figure 6: A simple Boltzmann machine

$$\{weights_{C_1}\}, \{weights_{C_2}\}, \{weights_{C_3}\},$$
$$\{h_1, x_1, o, x_1, x_2, x_3, x_4\}.$$

These components also correspond to the component subgraphs. The fourth and largest component subgraph when extended with its parents, as required for Definition 8, corresponds to the entire Figure 6(b) but with the directions of the arcs dropped. The cliques in this component subgraph with parents included are:

$$\{h_1, x_1, o, weights_{C_1}\}, \{h_1, x_2, o, weights_{C_2}\},$$
$$\{o, x_3, x_4, weights_{C_3}\}.$$

From Equations (3) and (4) it follows that the conditional probability is given by:

$$p(o, h_1 | x_1, x_2, x_3, x_4, weights_{C_1}, weights_{C_2}, weights_{C_3})$$
$$\propto \quad f_{C_1}(h_1, x_1, o, weights_{C_1}) \, f_{C_2}(h_1, x_2, o, weights_{C_2})$$
$$f_{C_3}(o, x_3, x_4, weights_{C_3}),$$

where the normalizing constant would be determined. Notice the terms $p(weights_{C_i})$ have been dropped from this expression because they normalize out. Since the variable $h_1$ is hidden, and the target is the probability of the output $o$, the conditional probability $p(o|x_1, x_2, x_3, x_4, weights)$ would be found via Bayes theorem.

$$p(o | x_1, x_2, x_3, x_4, weights_{C_1}, weights_{C_2}, weights_{C_3})$$
$$\propto \quad \sum_{h_1} f_{C_1}(h_1, x_1, o, weights_{C_1}) \, f_{C_2}(h_1, x_2, o, weights_{C_2})$$
$$f_{C_3}(o, x_3, x_4, weights_{C_3}),$$

where again the normalizing constant is to be determined. As this stands, this allows the functions $f_{C_1}$,

$f_{C_2}$ and $f_{C_3}$ to take on a general form so this is really a higher-order Boltzmann machine. Boltzmann machines traditionally only involve quadratic terms. The correspondence between stochastic neural networks and probabilistic networks is not exact.

### 6.3    Bayesian classifiers

Bayesian classifiers (Duda & Hart, 1973) are a broad family of supervised learning systems. Bayesian networks offer a rich representation for designing many different kinds of Bayesian classifiers, for instance illustrated with the Bayesian conditional trees of (Geiger, 1992). Chain graphs offer a richer family again of Bayesian classifiers, and a nice framework for their elicitation. During elicitation we can interpret the directed arcs in the "true"[1] chain graph as being causal connections, and undirected arcs as being associational connections. The language of chain graphs allow associations to be represented in a model in those situations where causality is perhaps difficult to interpret. This is illustrated by (Højsgaard & Thiesson, 1995). They present an example where a Bayesian classifier for *Coronary artery disease* is constructed by learning a chain graph from data, and then conditioning the chain graph for the key target variable *Coronary artery disease*, denoted $c$ using the formula

$$p(c|other\text{-}vars) = \frac{p(c, other\text{-}vars)}{\sum_c p(c, other\text{-}vars)},$$

where *other-vars* is the other variables in the graph. Prior information about the underlying "true" chain graph is elicited from a clinician. This prior information represents constraints on the eventual chain components, and directed and undirected arcs which definitely should or should not be present. The clinician's prior includes such constraints as (the corresponding variable in the graph follows in brackets):

- *sex* ($s$) is a causal factor for *smoking* ($S$),
- there is no association between *previous myocardial infarct* ($A$) and *angina pectoris* ($a$),
- there is possible association between ECG-examinations *Q-wave* ($Q$) and *T-wave* ($T$) but there is no causal link between the two.

A sample chain graph consistent with this prior knowledge is given in Figure 7 (this simplifies the situation presented in (Højsgaard & Thiesson, 1995)).

## 7    Chain graphs with plates

To represent data analysis problems within a network language such as chain graphs, some additions are needed. As a notational device to represent a sample—a group of like variables whose conditional distributions are independent and identical—plates are used on a chain graph (Buntine, 1994). By defining various operations on chain graphs with plates, such as conditioning and differentiation, useful algorithms can be pieced together for standard statistical procedures such as maximum likelihood or maximum *a posteriori* calculations, or the expectation maximization algorithm. Chain graphs with plates therefore represent a specification language for data analysis problems.

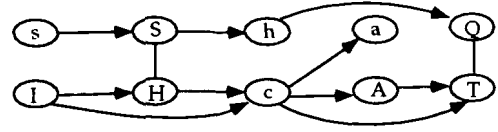

Figure 7: A chain graph for a Bayesian classifier to predict $c$

To introduce plates, consider the simplified version possible of a learning problem: there is a biased coin with an unknown bias for heads $\theta$. That is, the long-run frequency of getting heads for this coin on a fair toss is $\theta$. The coin is tossed $N$ times and each time the binary variable $heads_i$ is recorded. The graphical model for this is in Figure 8(a). The $heads_i$ nodes are shaded because their values are given, but the $\theta$ node is not. The $\theta$ node has a $Beta(1.5, 1.5)$ prior. The data for this problem is the sequence of heads variables. The key thing to notice is that for an "independently and identically distributed" sample, the network model for each case will be equivalent, and will be conditioned on the same model parameter, $\theta$. So the portion of the network corresponding to each i.i.d. case will be identical. Plates allow this duplication to be removed. The corresponding plate model for Figure 8(a) is given in Figure 8(b).

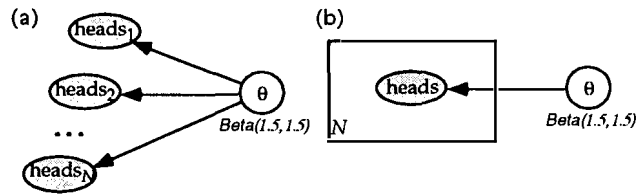

Figure 8: Tossing a coin: model without and with a plate

There can be multiple plates in the one diagram, indicating multiple tables in the data set. Data for the dollar-Deutsch mark exchange rate consists of a sequence of bids posted by banks. Original data takes the form of a date and time, the bid and asking price, and the bank code. This is converted into the mean bid-ask price, $(bid + ask)/2$, and the spread, $(ask - bid)$ because these variables more naturally representation the domain.

| | Date | Mean bid-ask | Spread | Bank |
|---|---|---|---|---|
| Sep 1 | 13:42:40 | 1.57395 | 0.0005 | CONY |
| Sep 1 | 13:42:45 | 1.5740 | 0.0010 | MGTX |
| Sep 1 | 13:43:14 | 1.57375 | 0.0005 | BBIX |

The skeptic would say that price differences are random and do not reflect any intelligent behavior by the banks. The skeptic therefore models this data as a random walk. It is clear that individual banks behave differently, so one way to do this is to have a separate random walk for each bank. This is shown in Figure 9(a) where each row in the data above corresponds to one instance of the plate. The variable *bid-ask-diff* is the difference in the mean bid-ask price from the bank's previous posting, which is random under a random walk model. Another model is to group banks into separate classes, where random walks now occur for the prices in each class of banks. Of course, we do not know what these classes are ahead of time, so this is a new kind of unsupervised learning problem. A probabilistic network for this second model is shown in Figure 9(b). In this model, the *bid-ask-diff* is assumed to be influence by the class, but is otherwise random. Notice in this model the outer plate is index by the bank, so each instance represents a different bank. The variable *Banks* is the number of banks. The inner plate is indexed by a single bank's prices, so each instance represents a different price for that bank. The variable $Ticks_{bank}$ is the number of prices posted for the bank. This represents another view of the data set, where we have a table of banks with an inner table of prices for each bank.

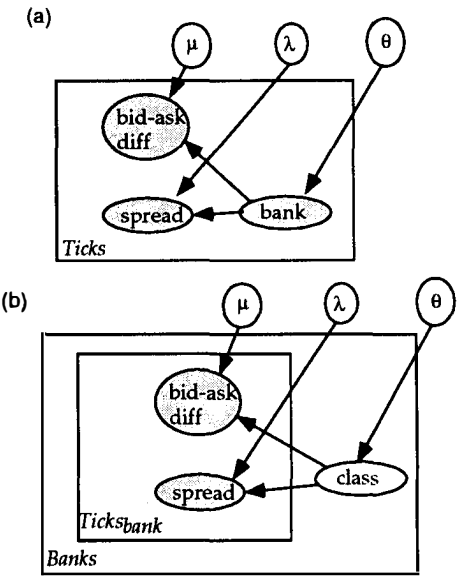

Figure 9: Two random walk models for banks.

**Bank**
**CONY**

| | Date | Mean bid-ask | Spread |
|---|---|---|---|
| Sep 1 | 13:42:40 | 1.57395 | 0.0005 |
| Sep 1 | 13:42:45 | 1.5740 | 0.0010 |

....
**BBIX**

| | Date | Mean bid-ask | Spread |
|---|---|---|---|
| Sep 1 | 13:43:14 | 1.57375 | 0.0005 |

....

The joint probability for the model of Figure 9(b) is given by

$$p\left(\theta, \mu, \lambda, \textit{bid-ask-diff}_{i,j}, spead_{i,j}, class_i : \right.$$
$$i \in Banks, j \in Prices(i)) =$$
$$p(\theta)\, p(\mu)\, p(\lambda) \prod_{i \in Banks} p(class_i | \theta) \prod_{j \in Prices(i)}$$
$$p(spread_{i,j} | \lambda, class_i)\, p(\textit{bid-ask-diff}_{i,j} | \mu, class_i) .$$

In this expression, $Banks$ is an index set for the banks and $Prices(i)$ is an index set for the prices for bank $i$. The product terms in the formula mirror the structure of the graph.

The notion of a plate is formalized below.

**Definition 11** *A chain graph $G$ with plates on variable set $X$ consists of a chain graph $G'$ on variables $X$ with additional boxes called plates placed around groups of variables. Directed arcs can only cross into plates, undirected arcs cannot cross plate boundaries, and plates can be overlapping. Each plate $P$ has an integer $N_P$ in the bottom left corner indicating its cardinality. The plate $P$ indexes the variables inside it with values $i = 1, \dots, N_P$. Each variable $V \in X$ occurs in some subset of the plates. Let indval$(V)$ denote the set of values for indices corresponding to these plates. That is, indval$(V)$ is the cross product of index sets $\{1, \dots, N_P\}$ for plates $P$ containing $V$.*

A graph with plates can be *expanded* to remove the plates and replace them with the contents duplicated. Figure 8(a) is an expanded form of Figure 8(b). This is done by duplicating the contents of the plate $N_P$ times, starting from the outermost plate and working inwards. The probability formula corresponding to a graph with plates can be written down from this expanded form, as was done for Figure 9(b) above. The details are tedious so I do not reproduce them here.

## 8 Conclusion

With chain graphs defined and operating for a large family of data analysis models in machine learning, neural networks and statistics, it is possible to crank the handle—apply the standard algorithmic methods for addressing probability problems—to create learning algorithm from the particular chain graph representation. Some examples of papers that discuss general methods for learning on graphical models are (Buntine, 1994; Gilks et al., 1993; Lauritzen, 1995).

### Acknowledgments

This paper was considerably improved by the suggestions from the reviewers.